\documentclass[biblatex,nonatbib]{article}

\usepackage[preprint]{nips_2018}

\usepackage[utf8]{inputenc} 
\usepackage[T1]{fontenc}    
\usepackage{hyperref}       
\usepackage{url}            
\usepackage{booktabs}       
\usepackage{amsfonts}       
\usepackage{nicefrac}       
\usepackage{microtype}      
\usepackage{graphicx}
\usepackage{caption}
\usepackage{subcaption}
\usepackage{amsmath} 
\usepackage{placeins}
\usepackage{diagbox}
\usepackage[backend=biber]{biblatex}
\title{Where does the Stimulus go? Deep Generative Model for Commercial Banking Deposits}

\author{
  Ni Zhan \\
  Carnegie Mellon University \\
  \texttt{nzhan@andrew.cmu.edu}
}

\begin{document}

\maketitle
\begin{abstract}
This paper examines deposits of individuals ("retail") and large companies ("wholesale") in the U.S. banking industry, and how these deposit types are impacted by macroeconomic factors, such as quantitative easing (QE). Actual data for deposits by holder are unavailable. We use a dataset on banks' financial information and probabilistic generative model to predict industry retail-wholesale deposit split from 2000 to 2020. Our model assumes account balances arise from separate retail and wholesale lognormal distributions and fit parameters of distributions by minimizing error between actual bank metrics and simulated metrics using the model's generative process. We use time-series regression to forward predict retail-wholesale deposits as function of loans, retail loans, and reserve balances at Fed banks. We find increase in reserves (representing QE) increases wholesale but not retail deposits, and increase in loans increase both wholesale and retail deposits evenly. The result shows that QE following the 2008 financial crisis benefited large companies more than average individuals, a relevant finding for economic decision making. In addition, this work benefits bank management strategy by providing forecasting capability for retail-wholesale deposits. 
\end{abstract}


\section{Introduction}



The Federal Reserve (Fed) has used quantitative easing (QE) as a response to economic slowdowns, including the 2008 financial crisis and recently Covid-19 crisis. In QE, the Fed purchases US bonds and other assets to increase money supply, improve liquidity, and encourage economic activity. The initial response to Covid-19 crisis, the Cares Act, injected \$2.2 Trillion into the US economy and saw a rapid, large increase in commercial banking deposits, which are money held at banks and the predominant part of macroeconomic M1 and M2 money supply measures. The scale and swiftness of deposit increase (19\% from March to May 2020) has been unprecedented in the past 20 years \cite{h6-fed}. With large amounts of QE and increase in money supply, we would like to examine the breakup of where the stimulus landed. In particular, this work focuses on determining how macroeconomic factors, including QE and loans, impact the financial position of large corporations and individuals in terms of their deposits at depository institutions. We focus on deposits because they are a large part of the money supply, and the perspective is useful for bank balance-sheet management and understanding where stimulus landed in the macroeconomy.

We classify deposits into two types: retail and wholesale. A deposit is considered "retail" when the depositor is an individual and "wholesale" when the depositor is a large company. Large companies (for example Apple, Walmart) have bank accounts at certain banks, but usually only at large banks. From a bank's perspective, deposits are balance-sheet liabilities, and the retail and wholesale deposits have different characteristics, such as liquidity outflow, risk exposures to interest rates, etc. \cite{banking-management-pwc}. This work aids managing banks' balance-sheet risks by increasing understanding of how macroeconomic factors, including QE, separately impact retail and wholesale deposits. From an economic policy perspective, this work examines the result of actions such as QE and increasing loan programs on deposits of individuals and corporations.

One of the main challenges is missing direct data for retail and wholesale deposits across the industry. Banks are unlikely to volunteer information about their client mix, and available data report total deposits irrespective of individual or corporation as the holder. Therefore this work has two main parts: first estimating an industry retail-wholesale deposit split, and then predicting retail-wholesale deposits' temporal changes based on macroeconomic factors. In the first part, we use the FDIC Statistics on Depository Institutions (SDI) dataset in an unsupervised machine learning problem. The SDI data has financial and demographic information about each depository institution, and we build a probabilistic generative model to infer retail-wholesale deposits from an institutions' data. In the second part, we build a time-series regression to predict industry retail-wholesale deposits using macroeconomic factors as input. To the authors' knowledge, this is the first work to estimate industry retail-wholesale deposits and examine their macroeconomic drivers.




\section{Related Work}

Theoretically in a closed economy, QE money will become apparent somewhere in the system, an idea known as flow-of-funds \cite{bain-1973-survey-applied-econom,green-2003-flow-funds}. Related work created a flow-of-funds simulation between banks, government, Fed, households, and firms \cite{caiani-2016-agent-based}, and our work also examines impact of these sectors on each other in an alternative method. Nagurney et. al. created a network flow-of-funds, and solved as optimization problem on case study with Fed balance sheet \cite{nagurney-1992-finan-flow}. These papers show the validity and opportunity of examining Fed and banks' balance sheet interactions, and our work also takes advantage of these theoretical underpinnings.

In addition, there is significant interest in QE impact on banking system, with studies of recent QE in the U.S. \cite{bnpparibas,rodnyansky-2017-effec-quant}, Euro-area \cite{horst-2020-impac-quant}, England \cite{joyce-2014-quant-easin}, and Japan \cite{matousek-2019-effec-quant-easin}. Matousek et. al. used vector autoregression (VAR) and found deposit growth after QE shock in large-sized banks with non-performing loans \cite{matousek-2019-effec-quant-easin}, and we likewise find deposit growth after QE. Other studies focused predominantly on QE impact on bank lending \cite{horst-2020-impac-quant,joyce-2014-quant-easin}, and bond prices \cite{christensen-2016-portf-model}. Our work is unique in examining QE impact on retail-wholesale deposits.

Some papers have examined bank types and retail-wholesale banks, although they defined retail-wholesale differently than we did \cite{gertler-2016-wholes-bankin,craig-2013-depos-market}. Gertler et. al. studied wholesale banks and their role in the 2008 financial crisis \cite{gertler-2016-wholes-bankin}. They defined wholesale bank as highly leveraged, funded by short term debt, such as overnight repo, and roughly corresponding with "shadow banks" such as Lehman Brothers. Craig et. al. examined liabilities structure with relation to competition among banks, and categorized liabilities into wholesale funding and deposits \cite{craig-2013-depos-market}. The difference is that our work categorized deposits by wholesale-retail, and did not examine liabilities other than deposits. Some papers clustered and differentiated banks using FDIC dataset \cite{cohen-2007-market-struc,adams-2007-who-compet}, similarly to this work. Cohen et. al. examined market share, impacts of product differentiation on bank competition and profitability, and classified banks into multimarket, community, and thrifts \cite{cohen-2007-market-struc}. Adams et. al. investigated market segmentation and substitution between thrift and multimarket banks, with implications for bank merger policy \cite{adams-2007-who-compet}.  

Other works that use deep learning for banking are summarized by review papers \cite{huang-2020-deep-learn, leo-2019-machin-learn}. In macroeconomics applications, Sevim et. al. used neural networks to predict currency crises \cite{sevim-2014-devel-early}. Other works focused on modeling banking default risk, credit risk, and liquidity risk \cite{huang-2020-deep-learn,leo-2019-machin-learn}. Banking risk management is an active area of interest \cite{craig-2013-depos-market,banking-management-pwc}, and our work also contributes to this topic.

Finally the generative model used in this work is inspired by variational autoencoders \cite{kingma2013auto, dayan1995helmholtz}, because our model also simultaneously draws samples from probability distributions and incorporates the samples into the loss function that trains the distributions.

\section{Methods}

\subsection{Dataset}

We used the FDIC SDI dataset, which contains financial information about each FDIC insured institution \cite{fdic-sdi}. The data is available in SQL style structure as CSV files, at quarterly frequency from 1993. (We use YYYY-Q\# to indicate end of quarter throughout.) There were around 5,300 institutions in 2019 and 10,100 institutions in 2000. Institutions include commercial banks, thrifts (such as former Washington Mutual), and savings banks. The FDIC classifies institutions by field "bank charter class" into six classes based on commercial, savings, thrift, and federal or state charter. We compared the deposits reported by SDI with H6 \cite{h6-fed} and H8 datasets \cite{h8-fed} published by the Fed. H8 large commercial aligns with FDIC commercial bank with national charter (bank class "N") to 0.4\% of domestic deposits from 2004 onwards. H6 commercial aligns with FDIC bank classes N, NM, SM to 1.1\% of domestic deposits from 2008 onwards. 


To provide additional intuition on retail-wholesale banks, we consider some examples. The largest banks JPMorgan Chase, Bank of America, Citigroup, and Wells Fargo have retail and wholesale deposits generally following their lines of businesses. For example, Bank of America's quarterly earnings supplement separates deposits by lines of businesses (LoBs): Consumer Banking, Global Wealth and Investment Management, Global Banking, and Global Markets. From description of Bank of America's LoBs, we can naively consider Consumer Banking as retail, Global Banking and Global Markets as wholesale, and Global Wealth and Investment Management as mix of retail-wholesale. However, obtaining the data for banks along LoBs is limited (depending on if the bank provides the information) and tedious (as it involves scraping quarterly PDFs). We also believe that some banks (Bank of New York Mellon, State Street) are purely wholesale, and many small banks serving local regions are purely retail.

The data on each bank includes balance sheet items, amounts for different types of loans, deposits in checkings, savings accounts, demographic information such as number of offices, and many other items. Because we want to estimate retail and wholesale deposits, which is unknown, we used data which intuitively may differentiate banks as having more wholesale or retail deposits. These useful data fields, which became inputs to our model, are described in Table \ref{table:bank-feat2} for 2019-Q4 snapshot. Banks with less than \$4.5M deposits or fewer than 250 accounts were dropped from subsequent analysis, and these banks had negligible effect on industry results. In Table \ref{table:bank-feat2}, "percent small amount" indicates percentage of deposits in accounts less than the FDIC insurance limit (\$100,000 before 2010 and \$250,000 after 2010). "Percent small accounts" indicates percentage of accounts smaller than the FDIC limit. To provide intuition, the first three features in Table \ref{table:bank-feat2} roughly measure bank size, and banks with large values for these features are more likely to have wholesale deposits. We also note the distributions of these features have very fat tails to the right, from examination of the log-values, particularly the means and maxes. For the last three features in Table \ref{table:bank-feat2}, we expect banks with larger values of these features to have larger composition of retail deposits, because small account balances are more likely retail and retail-dominated banks likely have more retail loans. Note that most banks have percent of small accounts very close to 98\%, while the percent of small amounts is much lower with mean 65\%. This indicates that account balances within a bank are distributed with fat tails on the right as well.

\newcolumntype{C}[1]{>{\centering\arraybackslash}m{#1}}

\newcolumntype{M}[1]{>{\centering\arraybackslash}p{#1}}
\newcommand\hd[2]{\multicolumn{2}{c}{\begin{tabular}{@{}c@{}}#2\end{tabular}}}

\begin{table}[!htbp]
\caption{Bank features and statistics for 5113 FDIC 2019-Q4 banks. Logs of dollar values were taken in thousands of dollars. Means and StdDev w.r.t. logs are parameters of a log-normal approximating the distribution of banks. Retail loans were calculated as sum of residential construction loans, residential real estate loans, and loans to individuals. Deposits and accounts have fat tails on the right. Larger \% small amount should indicate retail-dominated bank.}
  \label{table:bank-feat2}
  \centering
 \begin{tabular}{lccccccccc}
\toprule
 &
 \hd{2}{Domestic\\deposits}
&
\hd{2}{Deposits\\per office}
&
\hd{2}{Total\\accounts}
 & 
 \multicolumn{1}{C{1.5cm}}{\% small amount}
 & 
 \multicolumn{1}{C{1.5cm}}{\% small accounts}
 & 
 \multicolumn{1}{C{1.5cm}}{\% retail loans}
\\\cmidrule(r){2-3}\cmidrule(l){4-5}\cmidrule(l){6-7}
& \$B & Log   & \$B &Log   &  $\times 10^5$ &Log   \\
\midrule
Mean & 2.6  & 12.4 & 0.27 & 10.9 & 1.1 & 9.1 & 65 & 98 & 41\\
StdDev & - & 1.5 & - & 0.9 & - & 1.4 & 15 & 3 & 22    \\
Min &   0.006 & 8.7 & 0.002 & 7.7 & 0.003 & 5.5 & 0.1 & 46 & 0  \\
Max &   1400 & 21.1 & 123 & 18.6 & 980 & 18.4 & 100 & 100 & 100  \\
\bottomrule
\end{tabular}
\end{table}

\subsection{Deep Generative Model (DGM)}
\label{methods-dgm}

With the available data, we aim to estimate industry retail deposits. The problem is unsupervised because we do not have data for retail deposit fraction. We can think of retail deposit fraction as a hidden variable to be estimated from observed bank data. However using traditional methods such as Hidden Markov Model or Gaussian Mixture Model is difficult because each bank is not discretely retail or wholesale. Therefore we created a generative model customized for this problem. 

We consider the total deposits at a bank as the sum of account balances. If we knew the balance of each account, we would have the distribution of account balances. We do not have the complete distribution, however we have some information about the distribution of accounts at each bank from the SDI dataset, namely: number of small accounts, deposits in small accounts, total number of accounts, and total deposits. Intuitively, knowing the distribution of accounts helps determine retail-wholesale deposits because large accounts are more likely to be wholesale. In addition, each account is either retail or wholesale. We build this intuition into a probabilistic generative model.

Our model assumes account balances arise from two separate distributions: retail and wholesale, and we assume lognormal distributions for both. Lognormals work well for income data \cite{RePEc:wpa:wuwpmi:0505006} and seem reasonable given our fat tailed data. Each bank $b$ has five parameters at a point in time $t$: $p_{b,t}$ fraction of retail accounts, $\mu_{b,t,\text{ret}}$, $\sigma_{b,t,\text{ret}}$, $\mu_{b,t,\text{ws}}$, $\sigma_{b,t,\text{ws}}$ representing means and standard deviations of retail and wholesale distributions, respectively. Bank features in array form [\#banks, \#timesteps, \#features] are inputs to neural networks (NNs) which output the five parameters. The NNs provide some generalization by mapping banks with similar features to similar parameters, and we use NNs because the functional form between bank features and parameters is unknown. The NNs can also be regularized with, for example, L1 or L2 weight regularization, although we did not do this here. We then sample account balances from each parameterized distribution, i.e. $s_{b,t,\text{ret}} \sim \text{lognormal}(\mu_{b,t,\text{ret}}, \sigma_{b,t,\text{ret}}^2)$ and $s_{b,t,\text{ws}} \sim \text{lognormal}(\mu_{b,t,\text{ws}}, \sigma_{b,t,\text{ws}}^2)$. The objective we minimize includes error terms between observed data $\mathbf{x}$ and calculated metrics from the samples $\mathbf{v}$. The included observed data are total deposits, number of small accounts, and deposits in small accounts, denoted by set $\mathcal{I}$.
\begin{equation}
\mathcal{L}_{\text{samples}} = \frac{1}{n_t \cdot n_b}  \sum_t^{n_t} \sum_b^{n_b} \sum_{\mathcal{I}}  (\mathbf{x}_{b,t,i} - \mathbf{v}_{b,t,i})^2
\end{equation}
For the different metrics, $\mathbf{v}$ was calculated as follows.
\begin{equation}
\mathbf{v}_{b,t,\text{ total deposits}} = \frac{n_{\text{accounts}, b,t}}{n_{\text{samples}}}[p_{b,t} \sum s_{b,t,\text{ret}} + (1-p_{b,t}) \sum s_{b,t,\text{ws}}]
\end{equation}
\begin{equation}
\mathbf{v}_{b,t, \text{ deposits in small accounts}} = \frac{n_{\text{accounts}, b,t}}{n_{\text{samples}}}[p_{b,t}  \sum_{\mathcal{S}: s_{b,t,\text{ret}}< l} s_{b,t,\text{ret}} + (1-p_{b,t}) \sum_{\mathcal{S}: s_{b,t,\text{ws}}< l} s_{b,t,\text{ws}} ]
\end{equation}
\begin{equation}
\mathbf{v}_{b,t, \text{ number of small accounts}} = p_{b,t} P(s_{b,t,\text{ret}} < l) + (1-p_{b,t}) P(s_{b,t,\text{ws}} < l)
\end{equation}

where probabilities $P$ are given from the CDF, and $l$ is FDIC insurance limit.

Based on $\mathcal{L}_{\text{samples}}$, we see the model fits parameters by minimizing errors between actual bank metrics and simulated metrics from the generative process. The model's generative process is specifically formulated to allow estimation of retail and wholesale deposits. 
To perform inference, we calculate retail deposits $\mathbf{r}$ using \begin{equation}
\label{eq:retail-inference}
\mathbf{r}_{b,t} = \frac{n_{\text{accounts}, b,t}}{n_{\text{samples}}}p_{b,t} \sum s_{b,t,\text{ret}} \hspace{0.2cm}.
\end{equation} 

Because we assume wholesale distribution is distinct from and has larger mean than retail, we constrain the distribution parameters, or "give a prior", as shown in Eq. \ref{eq:loss-constrain}. 
\begin{equation}
\label{eq:loss-constrain}
\mathcal{L}_{\text{constrain}} = \lambda \sum_t^{n_t} \sum_b^{n_b} [ (\mu_{b,t,\text{ret}}- {\mu_{0}}_{t,\text{ret}})^2 + (\mu_{b,t,\text{ws}}- {\mu_{0}}_{t,\text{ws}})^2+(\sigma_{b,t,\text{ret}}- {\sigma_{0}}_{t,\text{ret}})^2+(\sigma_{b,t,\text{ws}}- {\sigma_{0}}_{t,\text{ws}})^2]
\end{equation}
where $\lambda$ is a hyperparameter balancing  $\mathcal{L}_{\text{constrain}}$ and $\mathcal{L}_{\text{samples}}$. Overall, we minimize $\mathcal{L}_{\text{constrain}}+\mathcal{L}_{\text{samples}}$. In Eq. \ref{eq:loss-constrain}, $\mu_{0}$ and $\sigma_{0}$ are our guesses, which vary over timesteps but do not vary over banks. In Section \ref{results-valid-dgm}, we validate the assumption that retail and wholesale distributions are separate.

The model was implemented in TensorFlow. We minimized the loss using stochastic gradient descent. We used one neural network for retail parameters, one for wholesale parameters, and one for fraction of retail accounts $p_{b,t}$. Each NN had one hidden layer with 50 reLU units, and retail and wholesale NNs had separate output layers for mean and standard deviation. NN inputs were features in Table \ref{table:bank-feat2}, specifically we used normalized log-values and fractional values instead of percentages. NN outputs for standard deviations were exponentiated to ensure positivity. NN output for $p_{b,t}$ was transformed using sigmoid to ensure it was between 0 and 1. The model was trained on data from 2000-Q1 to 2019-Q4, with priors for ${\mu_{0}}_{t,\text{ret}}$ linearly spaced from 1.2 to 1.5, ${\sigma_{0}}_{t,\text{ret}}$ from 1.3 to 1.5, ${\mu_{0}}_{t,\text{ws}}$ from 4.5 to 5.0, ${\sigma_{0}}_{t,\text{ws}}$ from 3.0 to 3.5, and $\lambda=0.05$. The priors were chosen based on results from Section \ref{results-valid-dgm}. We drew 10,000 samples from each retail and wholesale distribution during training and inference, and averaged 10 calculations for our final inference predictions (Eq. \ref{eq:retail-inference}) to decrease variability of sampling process. 

\subsection{Benchmark}

We compare our retail-wholesale estimation with a benchmark. The benchmark uses FDIC Branch Office Deposits dataset, which has deposits in each branch of each institution and is published annually from 1994 \cite{fdic-branch-office-dep}. For the benchmark, deposits in branches with less than \$500 Million deposits are considered all retail deposits, and deposits in branches with more than \$500 Million deposits are considered wholesale.

\subsection{Time Series Regression}
\label{methods-time-series}

We use the retail-wholesale deposits from the DGM as the ground truth, and predict their future movements based on macroeconomic data. If the time-series regression accurately predicts movements, then the model inputs are likely drivers, although this shows association but not  causation. The initial macroeconomic inputs considered were reserves, Treasury General Account (TGA), loans at commercial banks, retail loans at commercial banks, Fed assets, and currency. After trying different combinations of features, we used reserves, loans, and retail loans because they gave good predictions. Specifically, we used reserves (WRBWFRBL) and loans (TOTLLNSA) from FRED Economic Data \cite{fred-data}. Reserves indicates balances of bank deposits at the Fed. During QE, the Fed purchases assets with newly-created bank reserves, therefore amount of increase in reserves corresponds with degree of QE. Retail loans were from SDI dataset as indicated in Table \ref{table:bank-feat2}, although similar retail loan data is released weekly by Fed in H8 \cite{h8-fed}. 

Our model was a linear regression with four backward periods data as input. We found using less than four backward periods would decrease prediction accuracy, but greater than four periods did not significantly improve accuracy. We used quarter-over-quarter differences (q/q diff) for model inputs and outputs to achieve better numeric behavior and  stationarity. We used 55 prediction cases from 2003 to 2017 for training and 10 data points from 2017 to 2020 for testing. We provide an example of model prediction. We define q/q diff for 6-30-2020 to be level at 6-30-2020 minus level at 3-31-2020. If we were to predict deposit q/q diff for 6-30-2020, our input would be q/q diffs of data for 9-30-2019, 12-31-2019, 3-31-2020, and 6-27-2020. We use the most recent macroeconomic data available closest to the prediction date. Our model input data are released weekly by the Fed, so we estimate the most recent data could be three days prior to the prediction date. We used all zero biases in linear regression because we wanted to attribute all changes in deposits to the input features.



\section{Results}

\subsection{Validation of Retail and Wholesale Account Balance Distributions}
\label{results-valid-dgm}
We want to validate the assumption that retail and wholesale account balances arise from separate distributions. We fit two subsets of the banks (one representing all retail and one representing all wholesale) to simplified DGM with only one distribution. The parameters obtained with the retail subset of banks represent an estimated retail distribution, and parameters obtained with the wholesale subset represent an estimated wholesale distribution. We would like to observe a difference in parameters to confirm the assumption that retail and wholesale account balances can be modeled with two distinct distributions.

\begin{figure}[hbt!]
\centering
\begin{subfigure}{.5\textwidth}
  \centering
  \includegraphics[width=\linewidth]{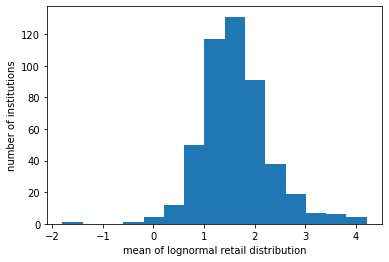}
\end{subfigure}%
\begin{subfigure}{.5\textwidth}
  \centering
  \includegraphics[width=\linewidth]{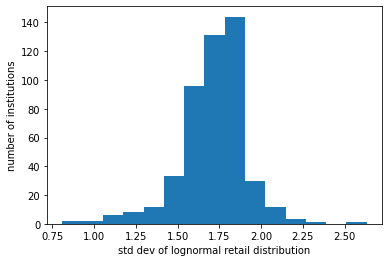}
\end{subfigure}
\caption{Means and standard deviations of lognormal distribution of account balances for thrift institutions in 2019-Q4}
\label{fig:thrift-dist}
\end{figure}

For the retail subset, we use thrift institutions (FDIC "SA" and "SB" bank class), and for the wholesale subset, we use State Street and Bank of New York Mellon (because we were unaware of other banks that are purely wholesale), and we perform the comparison for 2019-Q4. The means and standard deviations of the thrift institutions are shown in Fig. \ref{fig:thrift-dist}. Most means $\mu_{b,t,\text{ret}}$ are between 1.0 to 2.0. (The deposits are reported in thousands of dollars, therefore 1.0 represents \$$1000e$ (or \$2,718), and 2.0 represents \$$1000e^2$  (or \$7,389.)) Most standard deviations $\sigma_{b,t,\text{ret}}$ are below 2.0. For 2019-Q4, the best fit $\mu_{b,t,\text{ws}}$ for State Street and Bank of New York Mellon were 4.5 (\$54,500) and 3.5 (\$33,000), respectively, and the $\sigma_{b,t,\text{ws}}$ were 2.3 and 2.5, respectively. The distribution representing wholesale is shifted to the right of the retail distribution, which matches our assumption and intuition. We note that a small fraction of the thrift institutions had larger means, and it is possible that these were purely wholesale banks. For example, we found a thrift institution called NexBank which primarily serves institutional clients \cite{nexbank-thrift} and had mean 3.2 in our model. 

A deposit of a large company vs. high-net-worth individual may look the same to our model. However, we do use lognormals for both distributions, and sampling the lognormal retail distribution will result in large account balances attributed to high-net-worth individuals.  

\subsection{DGM Prediction}
We use the banks with at least one branch with over \$500 Million deposits in 2019 as training dataset, and fit the DGM neural network parameters with data from 2000-Q1 to 2019-Q4. We chose the training dataset this way because these banks are likely to have a mix of wholesale and retail deposits. The neural network input are bank features in array form [\#banks, \#timesteps, \#features], and the target are metrics in form [\#banks, \#timesteps, \#metrics], with metrics described in Section \ref{methods-dgm}. After objective minimization, we perform inference of retail deposits on all banks in FDIC SDI. 

We examined the predicted retail fraction for the larger individual banks. Some results of retail percent are listed here: Wells Fargo (55\%), JPMorgan Chase (50\%), Bank of America (64\%), Citi (35\%), Bank of New York Mellon (2\%), Ally (98\%), State Street (2\%), Morgan Stanley (38\%), Goldman Sachs (73\%). Overall the predictions seem reasonable, with Morgan Stanley and Goldman seeming high for being investment banks. The predictions are difficult to verify for most banks. In addition, the predictions seem to have high variance, and we took averages of 10 trials to smooth them. Further work can investigate how changes in training and model set-up affect prediction variance.

\begin{figure}[hbt!]
\centering
\begin{subfigure}{.5\textwidth}
  \centering
  \includegraphics[width=\linewidth]{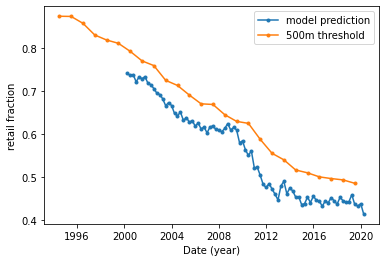}
  \caption{DGM prediction vs. 500M threshold benchmark}
  \label{fig:dgm-v-benchmark}
\end{subfigure}%
\begin{subfigure}{.5\textwidth}
  \centering
  \includegraphics[width=\linewidth]{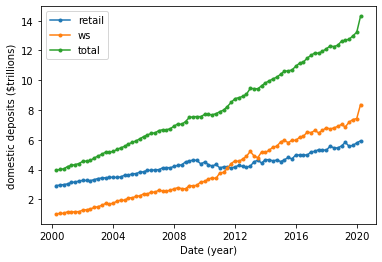}
  \caption{Nominal industry deposits 2000-2020}
  \label{fig:dgm-nominal}
\end{subfigure}
\caption{Deep Generative Model (DGM) predictions: fraction retail and nominal. DGM improves upon benchmark, and DGM predicts large jump in wholesale deposits for 2020-Q1.}
\label{fig:dgm-pred}
\end{figure}

Fig. \ref{fig:dgm-pred} shows the retail-wholesale deposit prediction for 2000 to 2020-Q1. Fig. \ref{fig:dgm-v-benchmark} shows the retail fraction over time for our model prediction vs. the 500M threshold benchmark. Both predictions show a downtrend in retail deposit fraction dropping from 70\% in 2000 to around 40\% in 2020. The trend may indicate that large companies' cash holdings grew faster than individuals' for the past twenty years. We note that it is possible the training dataset influenced the model prediction to have a similar trend as the 500M threshold method, however we did include all FDIC banks in model inference, and we believe sensitivity of model to training data can be explored in future work. We also assert that our model prediction improves upon the benchmark because our model gives four times as many predictions, that were estimated in a more precise, data-driven method. Our improved retail-wholesale estimation allows quantification of response to macroeconomic factors which is not possible with the rough yearly estimate from the benchmark. 

Fig \ref{fig:dgm-nominal} shows the same model prediction in a dollar amount format. The total industry deposits is readily available data, and our model contributes the split between retail and wholesale. The last timestep for 2020-Q1 was not in the training dataset, and our model predicted a large increase in wholesale rather than retail deposits. We believe the model accurately captured what actually occurred. Large companies were able to quickly respond to the shock of Covid-19 onset in the U.S. in March. Companies were able to draw loans from their agreements with banks, and the loans became deposits in the commercial banking system. On the other hand, it is likely that individuals did not have a large financial response to Covid-19 in March 2020. 

\FloatBarrier

\subsection{Time Series Model}
\begin{figure}[hbt!]
\centering
\begin{subfigure}{.5\textwidth}
  \centering
  \includegraphics[width=\linewidth]{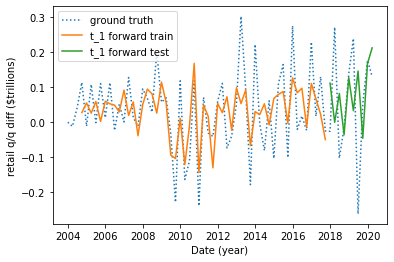}
  \caption{Retail deposits}
  \label{fig:ret-ts}
\end{subfigure}%
\begin{subfigure}{.5\textwidth}
  \centering
  \includegraphics[width=\linewidth]{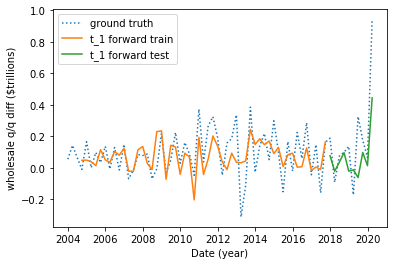}
  \caption{Wholesale deposits}
  \label{fig:ws-ts}
\end{subfigure}
\caption{Time series model forecast of ground truth deposits, matches well in direction and magnitude}
\label{fig:ts-pred}
\end{figure}

With the nominal prediction of retail-wholesale industry deposits, we build a time-series model with macroeconomic inputs. We take the DGM prediction as the ground truth for the time-series model, with model formulation as described in Section \ref{methods-time-series}. The model predicts one-step-forward quarter-over-quarter difference, and we compare the predictions against ground truth for retail and wholesale deposits in Fig. \ref{fig:ts-pred}. For both retail and wholesale deposits, the magnitude, variance, and direction of movements match the ground truth. This is true for train and test data, indicating that the model does not overfit. In addition we were particularly interested in model performance for test point 2020-Q1, and the model predicts a good amount of the jump in deposits then.

We now quantify the impact of macroeconomic inputs on the change in retail and wholesale deposits. The linear regression predicts forward-in-time changes and purposely has a zero bias so that deposit predictions are wholly accounted for by model input weights. We proxy the macro factor impacts by summing model weights for the four lag timesteps, and the impacts are shown in Table \ref{table:impacts}. To interpret the impacts, \$1.00 increase in reserves leads to around \$0.60 increase in wholesale and negligible change in retail deposits. The model also predicts approximately even change in retail and wholesale deposits as response to changes in loans. Because amount of increase in reserves corresponds with degree of QE, our results (namely the combined results from DGM and time-series models) predict that QE has a stronger increase on wholesale rather than retail deposits. This result matches our intuition that QE money is more easily captured by entities with already large funds and access to capital, such as large companies. 

\begin{figure}[hbt!]
\centering
\begin{subfigure}{.5\textwidth}
  \centering
  \includegraphics[width=\linewidth]{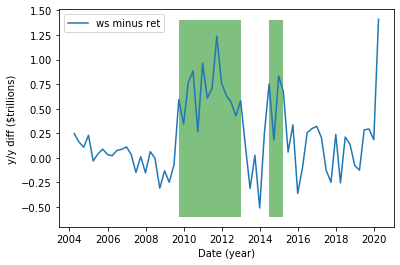}
  \caption{Wholesale minus retail year-over-year differences}
  \label{fig:ws-minus-ret}
\end{subfigure}%
\begin{subfigure}{.5\textwidth}
  \centering
  \includegraphics[width=\linewidth]{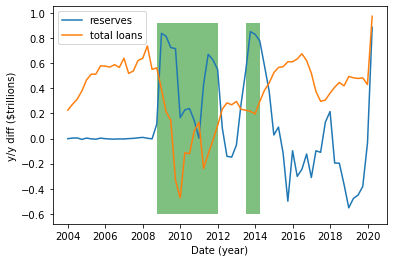}
  \caption{Reserves and total loans}
  \label{fig:input-lin-reg}
\end{subfigure}
\caption{Green time-periods represent faster than average wholesale deposit growth compared with retail. The corresponding model input time periods have 1-year backward lag, and showed large increase in reserves and negative to low loan growth.}
\label{fig:ts-historical}
\end{figure}

We examined the historical time periods when wholesale and retail deposit growth diverged. In the past 15 years, wholesale growth is generally faster than retail. Fig. \ref{fig:ws-minus-ret} highlights in green time-periods when wholesale was abnormally faster than retail deposit growth. These time-periods were 2009-Q3 to 2012-Q4 and 2014-Q2 to 2015-Q1. Fig. \ref{fig:input-lin-reg} highlights the model inputs for the same time-periods with a 1-year backward lag, because the time-series model essentially uses the previous year data for prediction. We observe large reserves growth with low loan growth; the time-periods were QE periods following the 2008 financial crisis. Our results indicate that QE contributed to wholesale deposits growing much faster than retail, historically. We note that QE time-period coincided with an economic recession, and the economic recession may have partially caused the low retail deposit growth. However this does not change the fact that wholesale deposits grew significantly faster in the same QE/recession time-period. If we have future rounds of stimulus for Covid, we will likely see the same pattern. The type of economic stimulus likely also matters. Stimulus that increases loan generation may more evenly benefit individuals and large companies. Future work can apply causal analysis to the associations between loans, QE and retail-wholesale deposits found here.






\begin{table}[!htbp]
\centering
\caption{Impact of Macroeconomic Inputs on Deposits}
\begin{tabular}{lrrr}
\toprule
\textbf{} & \textbf{Reserves} &  \textbf{Total Loans} & \textbf{Retail Loans} \\ \hline
Retail                & -6\% & 53\% & -23\%            \\
Wholesale           & 61\% & 64\% & -23\%             \\
Total              & 54\% & 116\% & -45\%             \\ 
 \bottomrule
\label{table:impacts}
\end{tabular}
\end{table}

\section{Conclusion}


This work developed the first model of U.S. banking industry deposits split by holder: individuals and large companies. We used a novel generative model which fit distributions of account balances per bank based on its financial data. The machine learning method took advantage of a rich FDIC dataset, which could be used in future investigations. With the wholesale-retail split, we showed that QE increases wholesale, and loans increase wholesale and retail deposits, a very intuitive result and what was observed following the 2008 financial crisis. The results have direct implication for economic policy, namely that QE has greater benefit for large companies and high-net-worth individuals. QE may increase economic inequality; at the same time other aspects of QE may backstop economic crises and can be explored in future work. Other extensions of this work include examining effects of interest rate on retail-wholesale deposits, and examining macro-factors impact on changes in market share across small regional banks and the largest banks.

\FloatBarrier
\newpage
\printbibliography

@article{bain-1973-survey-applied-econom,
  author =	 {A. D. Bain},
  title =	 {Surveys in Applied Economics: Flow of Funds
                  Analysis},
  journal =	 {The Economic Journal},
  volume =	 83,
  number =	 332,
  pages =	 1055,
  year =	 1973,
  doi =		 {10.2307/2230842},
  %url =		 {https://doi.org/10.2307/2230842},
  DATE_ADDED =	 {Thu Sep 10 12:48:45 2020},
}

@article{caiani-2016-agent-based,
  author =	 {Alessandro Caiani and Antoine Godin and Eugenio
                  Caverzasi and Mauro Gallegati and Stephen Kinsella
                  and Joseph E. Stiglitz},
  title =	 {Agent Based-Stock Flow Consistent Macroeconomics:
                  Towards a Benchmark Model},
  journal =	 {Journal of Economic Dynamics and Control},
  volume =	 69,
  number =	 {},
  pages =	 {375-408},
  year =	 2016,
  doi =		 {10.1016/j.jedc.2016.06.001},
  %url =		 {https://doi.org/10.1016/j.jedc.2016.06.001},
  DATE_ADDED =	 {Thu Sep 10 12:54:03 2020},
}

@article{green-2003-flow-funds,
  author =	 {Christopher J. Green and Victor Murinde},
  title =	 {Flow of Funds: Implications for Research on
                  Financial Sector Development and the Real Economy},
  journal =	 {Journal of International Development},
  volume =	 15,
  number =	 8,
  pages =	 {1015-1036},
  year =	 2003,
  doi =		 {10.1002/jid.961},
  %url =		 {https://doi.org/10.1002/jid.961},
  DATE_ADDED =	 {Thu Sep 10 12:55:12 2020},
}

@article{nagurney-1992-finan-flow,
  author =	 {Anna Nagurney and Merritt Hughes},
  title =	 {Financial Flow of Funds Networks},
  journal =	 {Networks},
  volume =	 22,
  number =	 2,
  pages =	 {145-161},
  year =	 1992,
  doi =		 {10.1002/net.3230220203},
 % url =		 {https://doi.org/10.1002/net.3230220203},
  DATE_ADDED =	 {Thu Sep 10 12:57:23 2020},
}

@inbook{gertler-2016-wholes-bankin,
  DATE_ADDED =	 {Thu Sep 10 13:25:59 2020},
  author =	 {M. Gertler and N. Kiyotaki and A. Prestipino},
  booktitle =	 {Handbook of Macroeconomics},
  doi =		 {10.1016/bs.hesmac.2016.03.009},
  pages =	 {1345-1425},
  publisher =	 {Elsevier},
  series =	 {Handbook of Macroeconomics},
  title =	 {Wholesale Banking and Bank Runs in Macroeconomic
                  Modeling of Financial Crises},
 % url =		 {https://doi.org/10.1016/bs.hesmac.2016.03.009},
  year =	 {2016},
}

@article{craig-2013-depos-market,
  author =	 {Ben R. Craig and Valeriya Dinger},
  title =	 {Deposit Market Competition, Wholesale Funding, and
                  Bank Risk},
  journal =	 {Journal of Banking \& Finance},
  volume =	 37,
  number =	 9,
  pages =	 {3605-3622},
  year =	 2013,
  doi =		 {10.1016/j.jbankfin.2013.05.010},
 % url =		 {https://doi.org/10.1016/j.jbankfin.2013.05.010},
  DATE_ADDED =	 {Thu Sep 10 16:41:14 2020},
}

@article{huang-2020-deep-learn,
  author =	 {Jian Huang and Junyi Chai and Stella Cho},
  title =	 {Deep Learning in Finance and Banking: a Literature
                  Review and Classification},
  journal =	 {Frontiers of Business Research in China},
  volume =	 14,
  number =	 1,
  pages =	 13,
  year =	 2020,
  doi =		 {10.1186/s11782-020-00082-6},
 % url =		 {https://doi.org/10.1186/s11782-020-00082-6},
  DATE_ADDED =	 {Thu Sep 10 16:57:50 2020},
}

@article{leo-2019-machin-learn,
  author =	 {Martin Leo and Suneel Sharma and K. Maddulety},
  title =	 {Machine Learning in Banking Risk Management: a
                  Literature Review},
  journal =	 {Risks},
  volume =	 7,
  number =	 1,
  pages =	 29,
  year =	 2019,
  doi =		 {10.3390/risks7010029},
 % url =		 {https://doi.org/10.3390/risks7010029},
  DATE_ADDED =	 {Thu Sep 10 17:02:34 2020},
}

@article{cohen-2007-market-struc,
  author =	 {Andrew M. Cohen and Michael J. Mazzeo},
  title =	 {Market Structure and Competition Among Retail
                  Depository Institutions},
  journal =	 {Review of Economics and Statistics},
  volume =	 89,
  number =	 1,
  pages =	 {60-74},
  year =	 2007,
  doi =		 {10.1162/rest.89.1.60},
%  url =		 {https://doi.org/10.1162/rest.89.1.60},
  DATE_ADDED =	 {Thu Sep 10 17:16:04 2020},
}

@article{adams-2007-who-compet,
  author =	 {Robert M. Adams and Kenneth P. Brevoort and
                  Elizabeth K. Kiser},
  title =	 {Who Competes With Whom? the Case of Depository
                  Institutions},
  journal =	 {Journal of Industrial Economics},
  volume =	 55,
  number =	 1,
  pages =	 {141-167},
  year =	 2007,
  doi =		 {10.1111/j.1467-6451.2007.00306.x},
 % url =		 {https://doi.org/10.1111/j.1467-6451.2007.00306.x},
  DATE_ADDED =	 {Thu Sep 10 17:24:23 2020},
}

@misc{h6-fed, url={http://www.federalreserve.gov/releases/h6/}, journal={Federal Reserve},year={2020},title={Federal Reserve H.6 Money Stock Measures}
}

@misc{h8-fed, url={http://www.federalreserve.gov/releases/h8/}, journal={Federal Reserve},year={2020},title={Federal Reserve H.8 Assets and Liabilities of Commercial Banks in the United States}
}

@misc{fdic-branch-office-dep, url={https://www7.fdic.gov/idasp/warp_download_all.asp}, journal={Federal Deposit Insurance Corporation},year={2020},title={Federal Deposit Insurance Corporation Branch Office Deposits}
}

@misc{fdic-sdi, url={https://www7.fdic.gov/sdi/download_large_list_outside.asp}, journal={Federal Deposit Insurance Corporation},year={2020},title={Federal Deposit Insurance Corporation Statistics of Depository Institutions}
}

@misc{fred-data, url={https://fred.stlouisfed.org/}, journal={FRED Economic Data},year={2020},institution={FRED Economic Data},title={FRED Economic Data}
}

@misc{nexbank-thrift, url={https://www.nexbank.com/}, journal={NexBank},year={2020},institution={NexBank},title={NexBank}
}

@misc{banking-management-pwc, url={https://www.pwc.com/gx/en/banking-capital-markets/assets/balance-sheet-management-benchmark-survey.pdf}, journal={PriceWaterhouseCoopers},year={2009},title={Balance sheet management benchmark survey},
}

@TechReport{RePEc:wpa:wuwpmi:0505006,
  author={Fabio Clementi and Mauro Gallegati},
  title={{Pareto's Law of Income Distribution: Evidence for Grermany, the United Kingdom, and the United States}},
  year=2005,
  month=May,
  institution={University Library of Munich, Germany},
  type={Microeconomics},
  url={https://ideas.repec.org/p/wpa/wuwpmi/0505006.html},
  number={0505006},
  doi={},
}

@article{matousek-2019-effec-quant-easin,
  author =	 {Roman Matousek and Stephanos Τ. Papadamou and
                  Aleksandar {\V{S}}evi{\'c} and Nickolaos
                  G. Tzeremes},
  title =	 {The Effectiveness of Quantitative Easing: Evidence
                  From Japan},
  journal =	 {Journal of International Money and Finance},
  volume =	 99,
  pages =	 102068,
  year =	 2019,
  doi =		 {10.1016/j.jimonfin.2019.102068},
 % url =		 {https://doi.org/10.1016/j.jimonfin.2019.102068},
  DATE_ADDED =	 {Sat Oct 3 22:52:18 2020},
}

@article{horst-2020-impac-quant,
  author =	 {Maximilian Horst and Ulrike Neyer},
  title =	 {The Impact of Quantitative Easing on Bank Loan
                  Supply and Monetary Policy Implementation in the
                  Euro Area},
  journal =	 {Review of Economics},
  volume =	 70,
  number =	 3,
  pages =	 {229-265},
  year =	 2020,
  doi =		 {10.1515/roe-2019-0033},
%  url =		 {https://doi.org/10.1515/roe-2019-0033},
  DATE_ADDED =	 {Sat Oct 3 22:59:32 2020},
}

@article{christensen-2016-portf-model,
  author =	 {Jens Henrik Eggert Christensen and Signe Krogstrup},
  title =	 {A Portfolio Model of Quantitative Easing},
  journal =	 {SSRN Electronic Journal},
  year =	 2016,
  doi =		 {10.2139/ssrn.2777690},
 % url =		 {https://doi.org/10.2139/ssrn.2777690},
  DATE_ADDED =	 {Sat Oct 3 23:16:33 2020},
}

@article{joyce-2014-quant-easin,
  author =	 {Michael Joyce and Marco Spaltro},
  title =	 {Quantitative Easing and Bank Lending: a Panel Data
                  Approach},
  journal =	 {SSRN Electronic Journal},
  year =	 2014,
  doi =		 {10.2139/ssrn.2487793},
%  url =		 {https://doi.org/10.2139/ssrn.2487793},
  DATE_ADDED =	 {Sat Oct 3 23:20:55 2020},
}

@article{rodnyansky-2017-effec-quant,
  author =	 {Alexander Rodnyansky and Olivier M. Darmouni},
  title =	 {The Effects of Quantitative Easing on Bank Lending
                  Behavior},
  journal =	 {The Review of Financial Studies},
  volume =	 30,
  number =	 11,
  pages =	 {3858-3887},
  year =	 2017,
  doi =		 {10.1093/rfs/hhx063},
 % url =		 {https://doi.org/10.1093/rfs/hhx063},
  DATE_ADDED =	 {Sat Oct 3 23:23:01 2020},
}

@article{bnpparibas,
  author =	 {Céline Choulet},
  title =	 {QE and bank balance sheets: the American experience},
  journal =	 {Conjoncture},
  volume =	 {},
  number =	 {},
  pages =	 {},
  year =	 2015,
  doi =		 {},
  url =		 {https://economic-research.bnpparibas.com/html/en-US/QE-bank-balance-sheets-American-experience-7/23/2015,25852},
  DATE_ADDED =	 {Sat Oct 3 23:23:01 2020},
}

@article{sevim-2014-devel-early,
  author =	 {Cuneyt Sevim and Asil Oztekin and Ozkan Bali and
                  Serkan Gumus and Erkam Guresen},
  title =	 {Developing an Early Warning System To Predict
                  Currency Crises},
  journal =	 {European Journal of Operational Research},
  volume =	 237,
  number =	 3,
  pages =	 {1095-1104},
  year =	 2014,
  doi =		 {10.1016/j.ejor.2014.02.047},
 % url =		 {https://doi.org/10.1016/j.ejor.2014.02.047},
  DATE_ADDED =	 {Sun Oct 4 15:24:59 2020},
}

@article{dayan1995helmholtz,
  title={The helmholtz machine},
  author={Dayan, Peter and Hinton, Geoffrey E and Neal, Radford M and Zemel, Richard S},
  journal={Neural computation},
  volume={7},
  number={5},
  pages={889--904},
  year={1995},
  publisher={MIT Press}
}

@article{kingma2013auto,
  title={Auto-encoding variational bayes},
  author={Kingma, Diederik P and Welling, Max},
  journal={arXiv preprint arXiv:1312.6114},
  year={2013}
  }

\end{document}